\pdfoutput=1
\documentclass[11pt, a4paper]{article}
\usepackage[utf8]{inputenc}
\usepackage[T1]{fontenc}
\usepackage{geometry}
\usepackage{amsmath, amssymb, amsfonts}
\usepackage{graphicx}
\usepackage{hyperref}
\usepackage{cite}
\usepackage{algorithm}
\usepackage{algpseudocode}
\usepackage{booktabs}
\usepackage{url}

\title{\textbf{The Meta-Prompting Protocol: Orchestrating LLMs via Adversarial Feedback Loops}}
\author{
    \textbf{Fanzhe Fu} \\
    Zhejiang University \\
    \texttt{ffanz@zju.edu.cn}
}
\date{\today}

\begin{document}

\maketitle

\begin{abstract}
The transition of Large Language Models (LLMs) from stochastic chat interfaces to reliable software components necessitates a fundamental re-engineering of interaction paradigms. Current methodologies, predominantly heuristic-based ``prompt engineering,'' fail to provide the deterministic guarantees required for mission-critical applications. We introduce the \textbf{Meta-Prompting Protocol}, a rigorous theoretical framework that formalizes the orchestration of LLMs as a programmable, self-optimizing system. Central to this protocol is the \textbf{Adversarial Trinity}, a tripartite topology comprising a Generator ($\mathcal{P}$), an Auditor ($\mathcal{A}$), and an Optimizer ($\mathcal{O}$). By treating natural language instructions as differentiable variables within a semantic computation graph and utilizing textual critiques as gradients , this architecture mitigates hallucination and prevents model collapse. We demonstrate the theoretical viability of this approach using declarative programming paradigms (DSPy) and automatic textual differentiation (TextGrad), establishing a foundation for ``Observable Software Engineering'' in the era of probabilistic computing.
\end{abstract}

\section{Introduction}

The deployment of Large Language Models (LLMs) in production environments has highlighted a critical disconnect between their probabilistic nature and the deterministic requirements of software engineering. As LLM parameter scales grow exponentially, we are witnessing a fundamental leap from Explicit Instruction-Based Programming to \textbf{Probabilistic Intent Orchestration} \cite{zhao2023survey}. While LLMs demonstrate emergent capabilities in code generation and logical reasoning, bridging the gap to production-grade software remains a significant engineering challenge due to the stochasticity inherent in autoregressive generation.

\begin{figure}[h]
    \centering
    \begin{minipage}[t]{0.44\textwidth}
        \centering
        \includegraphics[width=\linewidth]{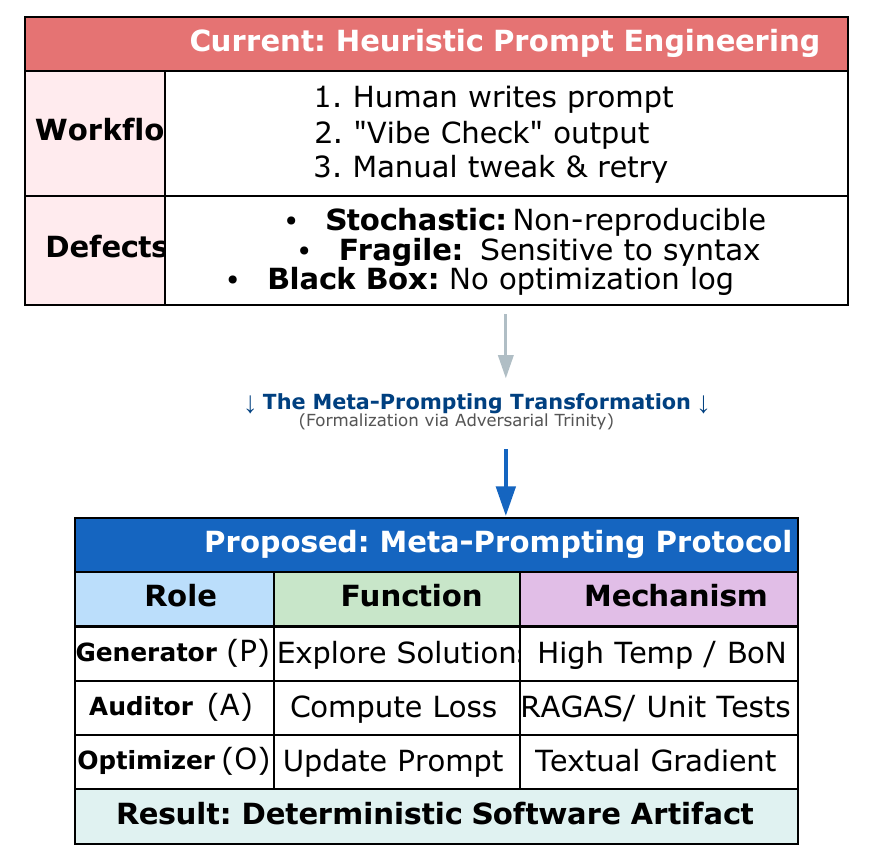}
        \caption{\textbf{Evolution of Interaction Paradigms.} The upper panel illustrates the prevailing Heuristic Prompt Engineering, characterized by cognitive bias and manual trial-and-error. The lower panel depicts the proposed Meta-Prompting Protocol, which transforms prompt construction into an automated, deterministic engineering closed-loop via the Adversarial Trinity architecture.}
        \label{fig:paradigm_shift}
    \end{minipage}
    \hfill 
    \begin{minipage}[t]{0.52\textwidth}
        \centering
        \includegraphics[width=\linewidth]{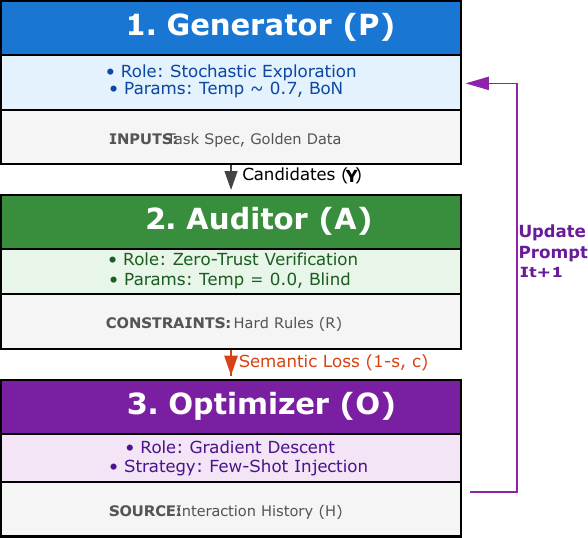}
        \caption{\textbf{Topology of the Adversarial Trinity.} The architecture comprises three functionally decoupled agents: the Generator ($\mathcal{P}$) for high-entropy exploration, the Auditor ($\mathcal{A}$) for zero-trust blind verification to compute semantic loss, and the Optimizer ($\mathcal{O}$) for executing backpropagation updates based on textual gradients. The solid loop signifies the recursive evolution of prompt parameters ($I_t$).}
        \label{fig:trinity_arch}
    \end{minipage}
\end{figure}

Traditional interaction models, often termed ``Prompt Engineering,'' remain in a heuristic, trial-and-error phase \cite{zhang2023metaprompting}. This ``black box'' approach introduces significant instability; minor perturbations in input syntax can lead to drastic fluctuations in output quality, making systems difficult to debug or optimize deterministically \cite{xi2023rise}. Figure~\ref{fig:paradigm_shift} illustrates this fundamental shift from manual, heuristic loops to the automated, deterministic engineering cycle proposed in this study. When migrating LLMs from auxiliary interfaces to the core logic of Autonomous Agents, reliability becomes the paramount constraint. The stochasticity of autoregressive generation conflicts with software engineering's requirement for correctness and reproducibility \cite{reynolds2021prompt}.


This report formally proposes the \textbf{Meta-Prompting Protocol} to address these limitations. We posit that prompts should not be viewed as natural language queries, but as \textbf{high-level source code}, where the LLM output is a transient compilation artifact. The remainder of this paper details the \textbf{Adversarial Trinity} architecture, defines the mathematical optimization rules for textual gradients \cite{pryzant2023protegi}, and analyzes the associated risks and mitigation strategies.

\section{Problem Formulation}

Formalizing the interaction with LLMs requires a shift from natural language heuristics to mathematical definition. This section defines the state space, the objective function, and the optimization constraints necessary to treat prompts as optimizing variables within a computational graph.

\subsection{Symbolic Definitions and System State}
Let $\mathcal{M}$ be a pre-trained LLM with frozen weights $\theta$. The system \textbf{State} is defined as a tuple $S = (\mathcal{I}, \mathcal{K}, \mathcal{H})$, where $\mathcal{I}$ is the Instruction Set (the Prompt), $\mathcal{K}$ is Contextual Knowledge, and $\mathcal{H}$ is the interaction history.

The \textbf{Generator Function} ($\mathcal{P}$) defines a stochastic mapping from input space $\mathcal{X}$ to output artifact space $\mathcal{Y}$:
\begin{equation}
y \sim \mathcal{P}(y \mid x, I, \mathcal{K}; \theta, \tau)
\end{equation}
Our objective is to find an optimal instruction set $I^*$ that maximizes the expected utility $U(y)$ over input distribution $\mathcal{D}$:
\begin{equation}
I^* = \mathop{\arg\max}_{I \in \mathcal{I}} \mathbb{E}_{x \sim \mathcal{D}} \left[ U(\mathcal{P}(y \mid x, I)) \right]
\end{equation}

\subsection{The Semantic Loss Function}
Since utility $U(y)$ is often non-convex and non-differentiable, we introduce the \textbf{Auditor Function} ($\mathcal{A}$) as an approximator. The Auditor evaluates artifact $y$ against hard constraints $\mathcal{R}$, outputting a scalar score $s$ and a structured \textbf{textual critique} $c$:
\begin{equation}
(s, c) = \mathcal{A}(y, \mathcal{R})
\end{equation}
We define the \textbf{Semantic Loss} $\mathcal{L}_{sem}$ as a composite vector acting as a \textbf{Textual Gradient} ($\nabla_{text}$):
\begin{equation}
\mathcal{L}_{sem} = (1 - s, c)
\end{equation}
This critique $c$ provides directionality in the semantic manifold, indicating how to minimize loss, a concept foundational to frameworks like TextGrad \cite{yuksekgonul2024textgrad}.

\subsection{Optimization Update Rule}
The \textbf{Optimizer Function} ($\mathcal{O}$) simulates stochastic gradient descent in the instruction space $\mathcal{I}$. In iteration $t$:
\begin{equation}
I_{t+1} = \mathcal{O}(I_t, c_t, \mathcal{H})
\end{equation}
This update reflects non-local, jump-based semantic refactoring, algorithmically realized by Bayesian optimization strategies in tools like DSPy \cite{khattab2024dspy}.

By establishing these definitions, we convert the subjective task of prompt engineering into a quantitative optimization problem. This allows the application of gradient-based methods in semantic space, providing a theoretical basis for automated system improvement.

\section{The Adversarial Trinity: Architectural Topology}

Operationalizing the mathematical formulation requires a specific architectural topology. We introduce the \textbf{Adversarial Trinity}, a system of three distinct agents—Generator, Auditor, and Optimizer—designed to decouple the phases of inference, verification, and refinement \cite{zhang2023metaprompting}. The topology of this architecture, detailing the cyclical data flow between the Generator, Auditor, and Optimizer, is depicted in Figure~\ref{fig:trinity_arch}.

\subsection{The Generator Agent ($\mathcal{P}$): Stochastic Exploration}
Agent P is the execution engine designed for high \textbf{Divergence}. It utilizes high-entropy sampling (Temperature $\tau \approx 0.7$) and Best-of-N (BoN) strategies to maximize coverage of the latent solution space \cite{kim2025cost}. Agent P operates on ``Task Specifications'' rather than restrictive step-by-step instructions, allowing for dynamic Chain-of-Thought (CoT) construction.

\subsection{The Auditor Agent ($\mathcal{A}$): Zero-Trust Deterministic Discrimination}
Agent A functions as the quality gate based on a \textbf{Zero Trust} assumption. To ensure stable gradient signals, Agent A uses \textbf{Blind Auditing} ($\tau = 0.0$), evaluating only the output $y$ against rules $\mathcal{R}$ without access to Agent P's internal reasoning. This prevents the ``Deceivable Discriminator'' phenomenon.

\subsection{The Optimizer Agent ($\mathcal{O}$): Meta-Cognitive Gradient Descent}
Agent O performs the ``backpropagation.'' It aggregates batch audit reports to identify systematic error patterns (e.g., ``80\% citation format errors'') rather than random noise \cite{chen2025optimizing}. Strategies include:
\begin{enumerate}
    \item \textbf{Constraint Hardening:} Explicitly adding negative constraints to the prompt.
    \item \textbf{Few-Shot Injection:} Compiling successful execution trajectories into in-context demonstrations \cite{khattab2024dspy}.
    \item \textbf{Strategy Refactoring:} If local refinements fail to converge, Agent O may alter the reasoning framework itself, such as switching from a Zero-Shot approach to ReAct or Plan-and-Solve architectures, effectively performing architectural search in the prompt space \cite{madaan2023selfrefine}.
\end{enumerate}

This tripartite separation of concerns ensures that the generation process remains creative while the verification process remains rigorous. This structural tension creates the necessary conditions for adversarial self-improvement, preventing the system from settling into local optima.

\section{The Iterative Loop Algorithm}

The protocol executes a recursive cybernetic loop to converge from a high-entropy state to a reliable low-entropy state. This section outlines the algorithmic control flow that governs the interaction between the agents defined in the Adversarial Trinity.

\begin{algorithm}
\caption{Meta-Prompting Iterative Optimization}
\begin{algorithmic}[1]
\State \textbf{Input:} Initial Prompt $I_0$, Rules $\mathcal{R}$, Training Data $\mathcal{D}_{train}$
\State $t \gets 0$
\While{$Score(\mathcal{A}) < Threshold$ \textbf{and} $t < MaxIterations$}
    \State \textbf{Batch Inference:}
    \State \quad Generate candidates $\mathbf{Y}_j$ for input batch $\mathcal{D}_{train}$ using Agent P
    \State \textbf{Audit \& Loss:}
    \State \quad Calculate critiques $(s_j, c_j) \gets \mathcal{A}(y_j, \mathcal{R})$
    \State \quad Aggregate gradients $\mathbf{C}_{agg} \gets \text{Cluster}(\{c_j \mid s_j < 1.0\})$
    \State \textbf{Optimization:}
    \State \quad $I_{t+1} \gets \mathcal{O}(I_t, \mathbf{C}_{agg})$
    \State \quad Inject self-corrected examples into $I_{t+1}$
    \State \textbf{Regression Testing:}
    \State \quad Validate $I_{t+1}$ on $\mathcal{D}_{gold}$ to prevent catastrophic forgetting
    \State $t \gets t + 1$
\EndWhile
\State \textbf{Return} Optimal Instruction Set $I^*$
\end{algorithmic}
\end{algorithm}

\subsection{Convergence Analysis}
The convergence of this protocol relies on the assumption that semantic space possesses gradients correlated with the utility function. While the space is discrete and non-convex, the \textbf{Batch Inference} step allows Agent O to smooth the loss landscape, filtering out stochastic failures (random noise) to focus on systematic biases \cite{yuksekgonul2024textgrad}. This mechanism is mathematically analogous to \textbf{Mini-batch Gradient Descent} in deep learning, where the optimizer updates parameters (prompts) based on the aggregate signal of a batch rather than individual unstable samples \cite{situmorang2025textualverifier}.

\section{Optimization Engines and Infrastructure}

The theoretical framework described above requires specialized infrastructure to manage state, track gradients, and execute recursive calls. This section surveys the computational frameworks, specifically DSPy and TextGrad, that enable the practical implementation of the protocol.

\subsection{DSPy: Declarative Self-Improving Pipelines}
DSPy abstracts prompts into \textbf{Signatures} (interfaces) and \textbf{Modules}. Its ``Teleprompter'' optimizers (e.g., MIPRO) automatically search the instruction and demonstration space, treating LLM calls as compile-able program modules \cite{khattab2024dspy}.

\subsection{TextGrad: Automatic Differentiation via Text}
TextGrad models the system as a computation graph where nodes are text variables. It backpropagates textual critiques from the Auditor to the Prompt variable, allowing for precise ``Credit Assignment'' in multi-agent systems \cite{yuksekgonul2024textgrad}.

\subsection{Observability}
Platforms like \textbf{LangSmith} provide deep tracing of CoT paths, while \textbf{PromptLayer} serves as a version control system for $I_t$, enabling the regression testing required in Algorithm 1 \cite{langsmith2024, promptlayer2024}.

Together, these tools form the ``compiler'' and ``debugger'' for the Meta-Prompting Protocol. They allow developers to move from manual string manipulation to declarative system design, where the optimization logic is handled by the infrastructure rather than the human engineer.


\section{Case Study: Automated Code Refactoring}

To demonstrate the efficacy of the Meta-Prompting Protocol, we consider a representative software engineering task: refactoring a legacy Python function to adhere to PEP-8 standards while optimizing time complexity.

\subsection{Execution Trace}
\begin{enumerate}
    \item \textbf{Initialization ($t=0$):} Agent P generates a solution that is syntactically correct but $O(n^2)$ complexity.
    \item \textbf{Audit ($t=0$):} Agent A detects the inefficiency. Semantic Loss $\mathcal{L}_{sem}$ contains the critique: ``Current implementation uses nested loops; optimized solution should target $O(n \log n)$.''
    \item \textbf{Optimization:} Agent O interprets the gradient and updates the prompt with a constraint hardening instruction: ``Explicitly use a hash map to reduce lookup time.''
    \item \textbf{Convergence ($t=1$):} Agent P regenerates an $O(n)$ solution. Agent A validates it against unit tests and clears the loop.
\end{enumerate}

This qualitative walkthrough exemplifies how the Adversarial Trinity navigates the solution space without human intervention.

\section{Evaluation Frameworks: The Mathematics of Audit}

The effectiveness of the Meta-Prompting Protocol is contingent upon the accuracy of the semantic loss function. This section details the quantitative frameworks used to transform subjective quality into objective, differentiable metrics suitable for the Auditor Agent.

\begin{itemize}
    \item \textbf{RAGAS (Reference-Free Evaluation):} Deconstructs quality into orthogonal metrics like \textbf{Faithfulness} (claims supported by context) and \textbf{Answer Relevance} (cosine similarity of embeddings) \cite{es2024ragas}.
    \item \textbf{DeepEval (Semantic Unit Testing):} Utilizes the \textbf{G-Eval} algorithm \cite{liu2023geval}, which calculates scores based on the expected probability of score tokens:
\end{itemize}
\begin{equation}
Score = \sum_{i=1}^{5} i \times P(score=i \mid y, \text{criteria})
\end{equation}

By employing these frameworks, the Auditor Agent provides a deterministic signal that guides the Optimizer towards the global maximum of the utility function.

\section{Systemic Risks: Model Collapse and Recursion}

The recursive nature of Meta-Prompting, where LLMs optimize other LLMs, introduces specific thermodynamic and informational risks. This section analyzes the phenomenon of Model Collapse and prescribes architectural safeguards .

\subsection{The Curse of Recursion}
Shumailov et al. define the ``Curse of Recursion,'' where training on self-generated data causes the model distribution to lose variance and converge to the mean \cite{shumailov2023curse}. In our context, relying solely on synthetic few-shot examples can lead to ``low-entropy'' states where the system loses the ability to handle edge cases \cite{shinn2023reflexion}.

\subsection{Mitigation Strategies}
\begin{itemize}
    \item \textbf{Golden Dataset Anchoring:} The optimization loop must include a mix (e.g., 20\%) of ground truth human-verified data to regularize the distribution.
    \item \textbf{Human-in-the-Loop Meta-Auditing:} Human engineers must review the \textit{changes} to the prompt suggested by Agent O, acting as a final safeguard against ethical drift or safety violations.
\end{itemize}

While Model Collapse represents a significant theoretical risk, it can be effectively managed through these strict architectural constraints, ensuring the system maintains sufficient entropy for robust generalization.

\section{Conclusion}

The transition from stochastic LLM interaction to reliable software engineering requires a rigorous theoretical foundation. This paper has presented the Meta-Prompting Protocol, a framework that formalizes prompts as source code and utilizes the Adversarial Trinity to perform Textual Gradient Descent.

By integrating declarative programming paradigms (DSPy), automatic differentiation (TextGrad), and rigorous evaluation (RAGAS), we have established a methodology for building observable, self-correcting AI systems. Although risks such as Model Collapse exist, the protocol's reliance on external ground truth anchors and human-in-the-loop meta-auditing provides a robust defense. Future work will focus on scaling this protocol to multi-agent swarms and formalizing the convergence bounds of semantic loss functions.

\bibliographystyle{ieeetr}
\bibliography{references} 

\end{document}